\documentclass{vgtc}                          

\ifpdf
  \pdfoutput=1\relax                   
  \usepackage{graphicx}                
  \DeclareGraphicsExtensions{.pdf,.png,.jpg,.jpeg} 
\else
  \usepackage{graphicx}                
  \DeclareGraphicsExtensions{.eps}     
\fi%

\graphicspath{{figures/}{pictures/}{images/}{./}} 

\usepackage{microtype}                 
\PassOptionsToPackage{warn}{textcomp}  
\usepackage{amssymb}
\usepackage{amsmath}
\usepackage{textcomp}                  
\usepackage{mathptmx}                  
\usepackage{times}                     
\usepackage{cite}                      
\usepackage{tabu}                      
\usepackage{booktabs}                  
\usepackage[super]{nth}
\usepackage{xcolor}

\onlineid{0}

\vgtccategory{Research}

\vgtcinsertpkg

\title{Visualizing RNN States with Predictive Semantic Encodings}

\author{Lindsey Sawatzky\thanks{e-mail: \href{mailto:lsawatzk@sfu.ca}{lsawatzk@sfu.ca}}
\and Steven Bergner\thanks{e-mail: \href{mailto:steven_bergner@sfu.ca}{steven\_bergner@sfu.ca}}
\and Fred Popowich\thanks{e-mail: \href{mailto:popowich@sfu.ca}{popowich@sfu.ca}}
}
\affiliation{\scriptsize School of Computing Science \\ Simon Fraser University}

\teaser{
  \centering
  \includegraphics[width=.95\linewidth]{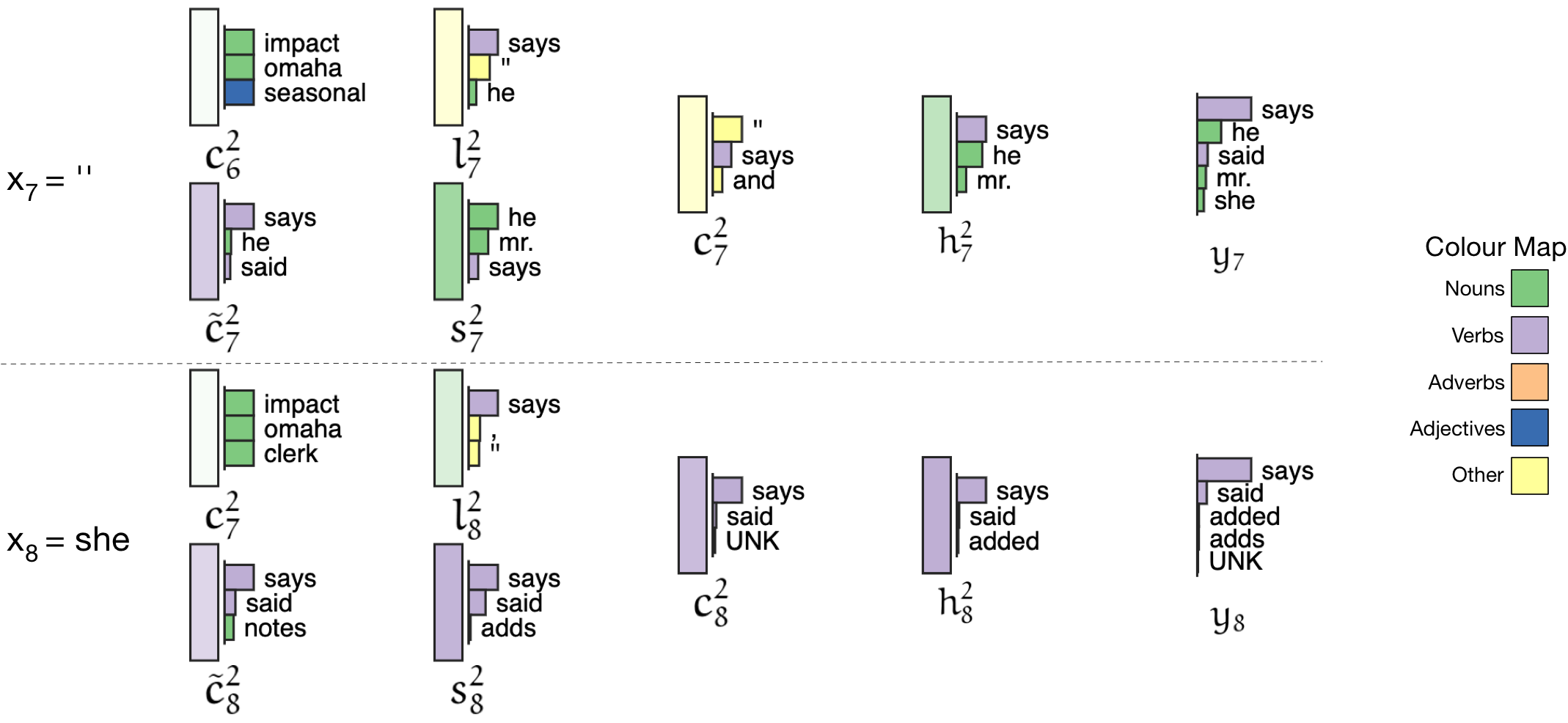}
  \caption{A subset of hidden state representations for the \nth{7} and \nth{8} timestep of the input sequence \textbf{\detokenize{``} \texttt{we stand in solidarity ,} \detokenize{''} \texttt{she emphasized .}}, discussed in detail in Section \ref{evaluation:1}.
  The semantics of hidden states are represented as their potential outputs, seen as colour rectangles and mini-bar charts.}
  \label{fig:teaser}
}

\abstract{
Recurrent Neural Networks are an effective and prevalent tool used to model sequential data such as natural language text.
However, their deep nature and massive number of parameters pose a challenge for those intending to study precisely how they work.
We present a visual technique that gives a high level intuition behind the semantics of the hidden states within Recurrent Neural Networks.
This semantic encoding allows for hidden states to be compared throughout the model independent of their internal details.
The proposed technique is displayed in a proof of concept visualization tool which is demonstrated to visualize the natural language processing task of language modelling.
} 

\CCScatlist{
  \CCScatTwelve{Human-centered computing}{Visu\-al\-iza\-tion}{Visu\-al\-iza\-tion techniques}{}
}

\begin{document}


\firstsection{Introduction}

\maketitle

Recurrent Neural Networks (RNN) are a machine learning technique which specializes in modelling sequential data due to their ability to capture and retain long term information.
Recent years have seen great success in applying these models to sequential data, such as language modelling~\cite{mikolov2010recurrent}, sentiment analysis~\cite{tang2015document}, and machine translation~\cite{sutskever2014sequence}.
However, despite their efficacy, the actual features and patterns which RNNs learn to model can be opaque, difficult to understand, and may even include unintended biases~\cite{bolukbasi2016man}.
This is a result of the black box nature of these systems which learn to model training data with little guidance about how to actually do so.

Visualizations are a powerful tool that can be employed to expose the inner details of these opaque models, taking a step towards better understanding and trusting their results~\cite{ribeiro2016should}.
However, understanding and visualizing RNNs comes with its own set of challenges.
One such challenge is that of interpreting the semantics of hidden states, vectors which encode information within the RNN.
These hidden states represent information in complex and highly non-linear ways, making them particularly difficult to interpret.

We aim to visualize some of the intuitions behind the internal representations RNNs learn to model.
We design a novel visual metaphor to address some of the challenges of visualizing RNNs, and in particular their hidden states.
This is Predictive Semantic Encodings (PSE) which are an interpretation of hidden state semantics that can be easily visualized and facilitate high level comparisons.
We develop a proof of concept interactive tool that uses this technique to visualize an overview of the Recurrent Neural Network.

\section{Related Work}\label{relatedwork}
Previous work exists in the field of visualizing RNNs, and in particular in terms of interpreting their hidden states.
One successful approach to this problem has been to interpret patterns of the hidden state activations with respect to the task objectives.
For example, Karpathy et al.~\cite{karpathy2015visualizing} show that some activations within the RNNs track long term syntactic language features, while Li et al.~\cite{li2015visualizing} use heat maps to understand the salience of activations and words with respect to the model output.
A similar heat map technique is used to show the ``attention'' a model pays to its sequential inputs at different timesteps across the recurrence relation~\cite{bahdanau2014neural, levy2018long}.

Although these techniques mainly focus on static visualizations, these and other works have been extended to incorporate interactive and complex visualizations.
LSTMVis~\cite{strobelt2018lstmvis} follows the approach of visualizing hidden state activations across time, allowing users to form and test hypothesis about the sequences evoking these patterns.
Rather than focusing on activation patterns across time, Ming et al.~\cite{ming2017understanding} relate hidden states to the model inputs and outputs through co-clustering and quantifying the model's expected response to these hidden states.
Kahng et al.~\cite{kahng2018cti} show and compare specific subsets of activations based off similar test instances.

More recently, Strobelt et al.~\cite{strobelt2019s} incorporate interpretation of hidden states within the context of visualizing a machine translation pipeline, allowing for users to debug multiple stages of the translation process.
Rather than interpret the RNN hidden states of a model with fixed parameters, Cashman et al. ~\cite{cashman2017rnnbow} instead focus on visualizing the training process itself by showing the gradient updates during the RNN learning process.

These studies all focus on visualizing single types of the RNN hidden states, such as their embeddings or memory cells.
They also use various techniques to relate the hidden states back to the inputs or outputs of the task.
However, to our knowledge previous work does not focus on representing the semantics of hidden states with respect to the task input/outputs in such a way that this interpretation may be visually compared across the various types of hidden states within the RNN.

\section{Background}\label{background}
A Recurrent Neural Network is a mathematical framework which maps a sequence of inputs $\{\mathbf{x}_1, .., \mathbf{x}_T\}, \mathbf{x} \in \mathbb{R}^V$ to a sequence of internal states $\{\mathbf{h}_1, .., \mathbf{h}_T\}, \mathbf{h} \in \mathbb{R}^N$.
Each $\mathbf{h}_t$ is updated by the non-linear function $\mathbf{h}_t \leftarrow RNN(\mathbf{x}_t, \mathbf{h}_{t-1})$.
This recurrent nature allows the previous hidden states $\mathbf{h}_{t-1}$ to capture the context necessary to compute the next RNN update.

Within the RNN formulation itself, any number of operations may be performed when leading from $\mathbf{x}_t$ and $\mathbf{x}_{t-1}$ to $\mathbf{h}_t$.
Typically, each of these operations is written as a vector definition, further describing the different kinds of hidden states comprising the model.
We refer to the concept of a vector definition, including that of the final hidden state $\mathbf{h}_t$ itself, as the ``kinds'' of hidden states of the RNN.
Each kind of hidden state describes a specific component of the recurrent model.

Depending on the specific task against which the RNN is deployed, the final hidden state of a timestep is fed into a function $F(\mathbf{h})$ to produce an output $y$.
In the case of a classification problem, $\mathbf{y} \in [0, 1]^K$ represents a probability distribution across $K$ labels.

This study looks specifically at the Long Short-Term Memory architecture (LSTM)~\cite{hochreiter1997long}, a specific type of RNN.
The detailed mathematical notation of the LSTM is laid out in the Appendix.

\section{Design}\label{design}
We set out to design a visualization technique to help researchers better understand the general intuitions RNNs capture.
Our study follows the nested model of visual design and validation described by Munzner~\cite{munzner2009nested} which lays out four layers, each built off the one before.

\subsection{Layers 1 \& 2: Domain \& Abstraction}
We preface the design by first describing the user groups interested in studying RNNs at this level of detail.

Strobelt et al.~\cite{strobelt2018lstmvis} develop three high level categories of user groups interested in RNNs; Architects, Trainers, and End Users.
We look specifically at users from the various groups with some degree of pre-existing understanding of neural networks, primarily Architects and Trainers.
These users want to understand the detailed behaviours RNNs learn to model, whether that understanding may be used to drive development of architectural improvements or to gain confidence that their results generalize beyond the training data.
Moreover, these users want an intuition for what the RNN represents in order to better grasp how it functions.

From this group of users elucidate a series of questions which they are interested in asking.
1) What information flows and changes from the inputs through to the outputs of the RNN?
2) Where are changes more and less prominent?
3) Does our intuition about the role of the mathematical components match that of the actual RNN behaviour?
At a high level, all these questions seek to understand how the model behaves from a mathematical/computational perspective.
From these questions, we ascertain abstract operations and data types which can later be mapped to visualization techniques.

Vectors are the most natural concept to focus on for study, not simply because there is precedent in previous works~\cite{li2016understanding, kahng2018cti, strobelt2018lstmvis}, but mainly because the RNN description uses vectors as the core element of notation.
Focusing on vectors ties directly to how these users naturally describe these models, ensuring visualization accessibility.

Implicit within the concept of RNN hidden state vectors are the dimensional values they encode, typically referred to as activations.
These activations are the focus of much of the previous work in visualizing RNN details.
However, in the context of visualizing high level information flow, it is dubious what information may be gleaned by observing the activation values themselves.
That is, visualizing these numeric values without a relatable interpretation of their meaning may not provide insights into the model behaviour.
Therefore, for the purposes of this work which intends to explore the flow of information throughout the RNN, we choose to abstract away activation details into the concept of a hidden state as a whole.

With this data type in mind, we explore the options for visualization operations that may be performed on hidden state vectors.
Since users want to interpret hidden states as a whole, we propose hidden state comparisons as the primary visualization operation.
By comparing the change in hidden states along the components of the RNN, these types of high level questions about the flow of information may begin to be answered.


\subsection{Layers 3 \& 4: Encoding-Interaction \& Algorithm}
Continuing with the final two layers of the nested model of visual design, this section describes the development of a visual encoding and algorithms to address requirements from the previous section.

We recall from the previous section the desire to express the intuitions encoded within RNN hidden states as a whole.
To do so, we introduce the semantics of hidden states as a probability distribution over the task output labels.

More formally, we consider the context free\footnote{The function is context free in that it makes a prediction similar to that of the RNN, but without any surrounding hidden state context.} function $G(\gamma, \mathbf{v})$ which produces a probability distribution over the outputs $\mathbf{y}$, where $\gamma$ denotes the specific kind of hidden state of the instance $\mathbf{v}$.
Indeed, this formulation is analogous to the classification task of the RNN itself, and the function $F(\mathbf{h})$ can be seen as a specialization of $G$ such that $F(\mathbf{h}) = G(\text{final-hidden-state}, \mathbf{h})$.

The function $G$ is modelled separately from that of the actual RNN and its classification function $F$, allowing it to represent arbitrary hidden states.
$G$ is trained on the hidden state instances derived from the sequential dataset once training the RNN is complete so as to capture the final semantics learned by the model.

Although the task of the RNN is to specifically predict the output labels, the semantic we propose may similarly be used to encode hidden states with respect to the input labels they correspond to.
It also may be used to predict the output labels at varying timesteps of the RNN, so that differing layers of semantics may be expressed by any particular hidden state.
The choice of which semantic(s) to use will depend on the specific visualization task.

Notice, the notion of relating hidden states in some way to the task inputs or outputs is not a new one, as for example is done in RNNVis~\cite{ming2017understanding}.
However, to our knowledge other work has not drawn a direct parallel between hidden states encoding a probability distribution similar to that of the task itself.

The advantage to this semantic formulation is that hidden states can share the same visual encoding as that of the RNN output prediction.
Moreover, this formulation describes hidden states in a way that is invariant of their internal significance.
These attributes allow users to seamlessly transition between comparisons of hidden state semantics amongst the components of the RNN as well as the task itself.

With these semantics in place, we develop a simple visual encoding designed to represent probability distributions across typical outputs of RNNs.
We use a mini-bar chart where bars represent class labels and the bar length represents the probability magnitude.
Since there may be many class labels, only the top-$k$ probabilities are shown in the bar chart, where $k$ is selected depending on the visualization context.
The class labels are shown in descending order based off the probability distribution so that the mini-bar chart represents the most likely outcomes.
Finally, each bar is coloured according to a user specified colour mapping, e.g. to show information about parts of speech.
This allows for designers to draw attention to class labels as applicable to the visualization task.

We augment the mini-bar chart with a redundant encoding that describes the probability distribution more holistically and facilitates a quick, high level comparison of the hidden state semantic.
The redundant encoding uses a simple coloured rectangle, where the colour also comes from the visualization's colour mapping.
The colour is produced by first selecting which mapped colour holds the largest sum of probabilities in the top-$k$ predictions from the distribution.
Then, a point is interpolated inversely proportional between this most likely colour and the colour ``white''.

\begin{figure}
 \centering
 \textbf{PSE Visual Encoding}\par\medskip
 \includegraphics[scale=.5]{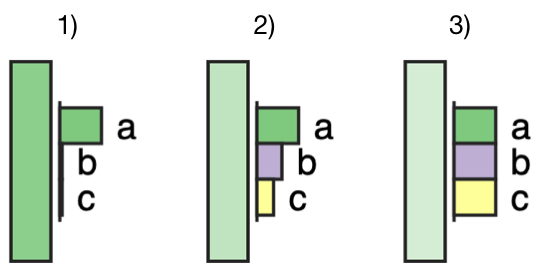}
 \caption{Visual encoding for the PSE with various probability distributions.
 In each case, the dominant colour of the probability distribution is green to the degrees of 1) 90\%, 2) 50\%, and 3) 33.33\%.}
 \label{fig:visualencoding}
\end{figure}

By example, if the mass of the probability distribution describes predictions all mapped to the same colour, then the coloured rectangle will reflect this colour without any white dilution.
However, if the mass of the probability distribution is roughly uniform across the colour mapping, then the most likely colour will be diluted to almost completely white.
In this way, the colour interpolation encodes the degree by which the probability distribution represents the dominant class labels.
Figure \ref{fig:visualencoding} shows examples of this visual encoding.

\subsection{Interactive Visualization Framework}
We implement a simple proof of concept tool focused on observing the RNN hidden states and architecture as a whole.

As comparing hidden states is important for understanding the flow of information throughout the model, the visualization shows many of the different kinds of hidden states comprising the model.
These are arranged according to their relative position in terms of the progression of time $t$ in the sequential input data, where time is represented as progressing from top to bottom in the display.
Additionally, the hidden states are arranged relative to their dependence in the RNN definition, so that those more directly dependent on the inputs $\mathbf{x}_t$ are positioned to the left of the visualization, while those closer to the outputs $\mathbf{y}_t$ are placed on the right.

The visualization tool allows for users to provide arbitrary sequential inputs to observe and compare the elicited hidden state representations.
It should be noted that this high level view of the RNN architecture is particularly suited for showing the intuition and conceptual interaction behind the detailed components of the model.

\section{Evaluation}\label{evaluation}
We evaluate our design with the following set of studies.
Indeed, these are preliminary results that encourage further follow up.

Although the proposed design element is applicable to any sequential modelling task, we focus on the well known language model task.
A language model is a natural language processing problem whereby each input word must produce a probability distribution describing the likelihood of the words that follow it.

We evaluate these studies on the Penn Treebank~\cite{marcus1993building} language model dataset, trained on a $300 \times 2$ LSTM.
Numerals and infrequent words are mapped to the \textbf{NUM} and \textbf{UNK} tokens, so that the resulting vocabulary consists of 14787 unique words.
The dataset contains 54020 and 6754 training and testing sentences, and the model is trained to achieve a perplexity of $66.5$ on the held out test set.
For the visualization, we use a colour mapping based on coarse level part of speech tags to highlight grammatical switches in the language model.
The colour interpolation for the PSE sets the top-$k$ as 10\% of the vocabulary.

\subsection{Case Study: Visualizing Information Flow}\label{evaluation:1}
In this case study, we explore the high level flow of information through the RNN.
The user chooses the input \textbf{\detokenize{``} \texttt{we stand in solidarity ,} \detokenize{''} \texttt{she emphasized .}} to observe the transition from the direct quote to end of the sentence.
The full visualization for this input and other examples can be found externally~\cite{Sawatzky_2019}, while Figure \ref{fig:teaser} shows just hidden states from the final layer of LSTM for the \nth{7} and \nth{8} timesteps.

The language model, who's outputs are visualized under the $\textbf{y}_t$ labels in the figure, has learned to immediately predict verbs or nouns to follow the end of quotation sequence.
By looking at the hidden state PSEs for the top row, although some of the colour interpolations agree with this semantic, it appears that the RNN is representing some degree of uncertainty at this stage in the state it models.
This uncertainty is manifested two fold, firstly by the weakly coloured purple and green interpolations, and secondly by the $\mathbf{l}_7^2$ and $\mathbf{c}_7^2$ hidden states instead forming a yellow-based interpolation.

Interestingly, the bottom row corresponding to the input of the word \textbf{she} appears to represent much more certainty in its prediction of a verb to follow.
This is an indication that the hidden states of the RNN at this point are generally confident about what will follow in the language model.
Since the input at this point is a noun, at a high level this explains how the internal state of the RNN has resolved the uncertainty from the previous timestep.

Focusing more closely on these PSE details, other aspects of the RNN internal behaviours are revealed.
In both rows of the figure, the previous Cell State $\mathbf{c}_6^2$ and $\mathbf{c}_7^2$ has interpolated a mainly white colour, indicating how the PSE for these hidden states represent a distribution without any particularly dominant label grouping.
This roughly uniform distribution can be further observed by looking at the top-3 predicted classes, which all sit flat together with equal probabilities.
By conveying this information, the PSE visual encoding shows how the previous Cell States must be gated to formulate the Long-term Memory before they can indicate a meaningful prediction.

Another interesting example is in $\mathbf{l}_8^2$, which indicates a probability distribution generally predicting green nouns, despite the fact that the input at this timestep should have switched the semantics to a purple verb prediction.
Although this PSE interpolation is a faded green colour, it shows how the Long-term Memory is still retaining semantics which indicate a noun prediction.
However, when this hidden state is combined with the Short-term Memory of $\mathbf{s}_8^2$ to produce $\mathbf{c}_8^2$ we see how the internal representation of the LSTM has switched to indicating a verb.
This reveals how the internal mechanics of the LSTM at this point in the model are combined to make a prediction, with the semantics from the Short-term Memory passing through as a result of that interaction.

Notice this effect is not predetermined, as seen for example across the same interaction in the previous timestep.
Instead there, it appears that the semantics of the Long-term Memory $\mathbf{l}_7^2$ win out when building $\mathbf{c}_7^2$.
At that point, it appears to be the Output Gate (not visualized) which switches out the indication of a yellow-function word to produce the final hidden state $\mathbf{h}_7^2$ predicting a noun/verb to follow in the model.

Observing the RNN model at this level of detail begins to reveal important nuances to the decisions its makes.
Moreover, these nuances are made available particularly through the visualization and comparison of the different hidden states comprising the model.

\subsection{Accuracy of Predictive Semantic Encodings}\label{evaluation:2}
This experiment investigates the accuracy of the Predictive Semantic Encoding model formulation.
Additionally, we use this formulation as a means by which the hidden states of the RNN can be understood.

In order to draw parallels with the RNN, we train and test on the full dataset of hidden states derived from the original sequential training and testing data.
This results in a total of 20M training and 2.6M testing data pairs across the 17 various LSTM hidden states (1 word embedding + 8 hidden states across 2 layers).

The PSE is evaluated using perplexity, which is the same metric as that of evaluating the language model task.
\begin{align}
Perplexity(X) & = \sqrt[T]{\prod_{t=1}^T \dfrac{1}{P(x_t|w_1...w_{t - 1})}}
\label{equation:perplexity}
\end{align}
With $X$ representing the sentence input and $P(x_t|x_t...x_{t - 1})$ as the model's conditional probability of a word $x_t$ given the context of previous words $x_t...x_{t - 1}$.
As perplexity is normalized over the length $T$ of the input, it is suited for comparisons between the sequential RNN data and that of the non-sequential PSE.

\begin{figure}
 \centering
 \textbf{PSE Accuracy per Hidden State}\par\medskip
 \includegraphics[scale=.35]{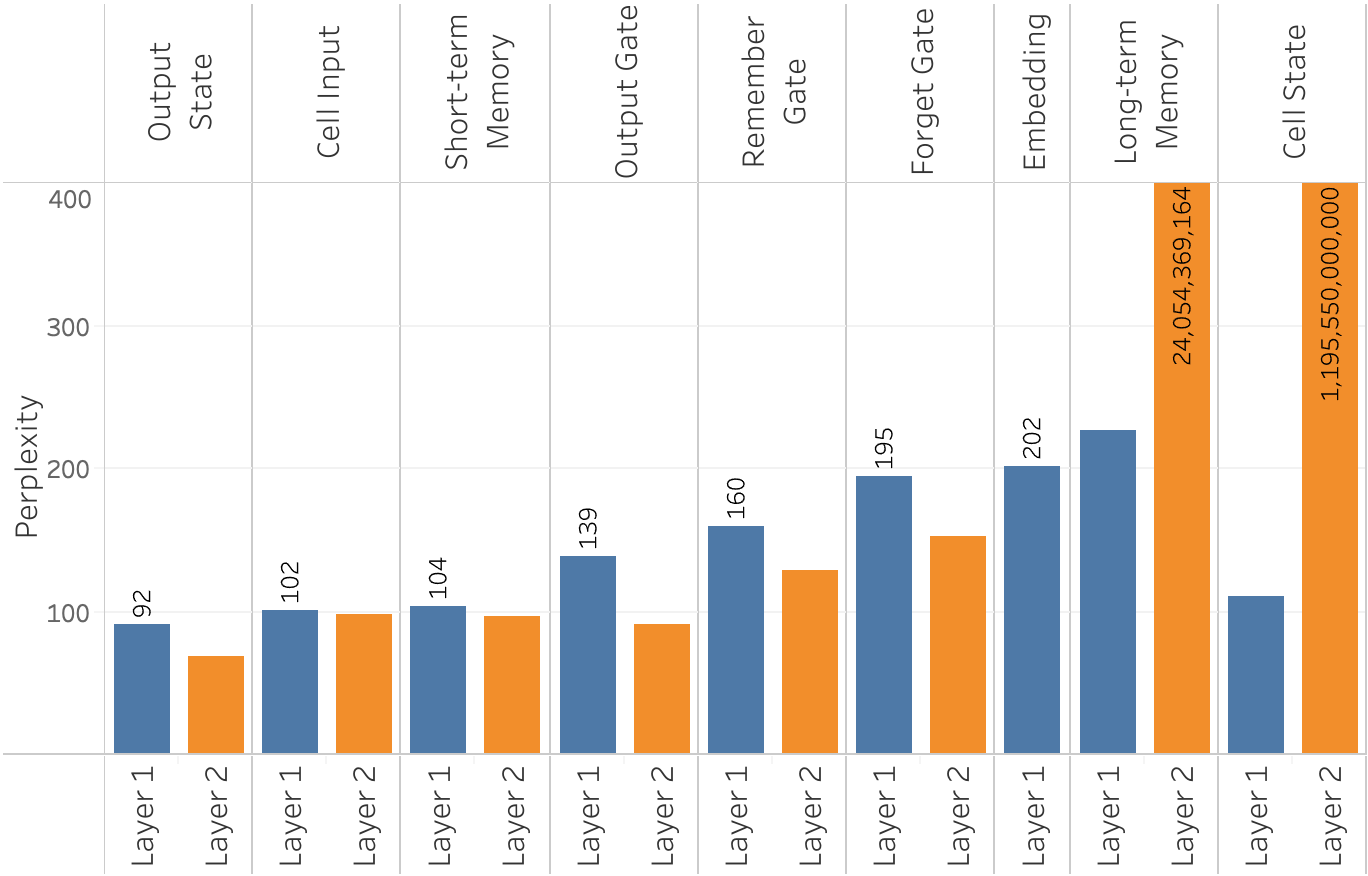}
 \caption{Per hidden state test set perplexity of the fully trained PSE using a linear classifier.
 Note the y-axis is truncated at $400$, despite some values exceeding this range.}
 \label{fig:pse}
\end{figure}

Figure \ref{fig:pse} shows the results of this experiment using a simple linear classifier as the PSE function $G$.
Notice, $G$ is modelled with a separate set of parameters for each kind of hidden state.

Firstly, we ensure parity with the RNN by examining the \nth{2} layer Output hidden state performance.
This hidden state achieves a test set perplexity of $70$, the best of all the hidden states, and critically this value is on par with the RNN which achieved a perplexity score of $66.5$.
This result establishes that the PSE are an accurate generalization of the RNN classification function $F$.

Next, the result that immediately stands out is at the \nth{2} layer for the Long-term Memory and Cell hidden states.
These values are several orders of magnitude larger than any other hidden states.
Despite this startling result, this finding actually explains an important aspect to the RNN architecture.
Specifically, the Long-term Memory and Cell States represent information that spans through the LSTM across time, rather than only containing information relevant to the current timestep.
As a result, these hidden states, especially at the final layer of the LSTM, are not reliable indicators of what final prediction will be made by the language model.

Another way to think about this is that the information from these hidden states is highly dependent on other pieces of context represented throughout the LSTM in making a final decision at any given timestep.
Therefore, it is not surprising that the Long-term Memory and Cell behave so poorly for this task.

We observe that the best performing hidden state is the Output of the final layer.
This is what we expect, since the model for this hidden state is equivalent to that of the actual RNNs classifier $F$.
Notice, the only difference in the LSTM architecture between the Output and the Cell State, which contrast the best and worst case results in the figure, is the Output Gate.
This fact highlights the important role this particular gating unit plays in the RNN.


It is worth noting that although these several kinds of hidden states are poor estimators of the RNN itself, that does not mean they perform poorly as Predictive Semantic Encodings.
Rather, this finding suggests they are working correctly, since the idea of a PSE is not to mimic the RNN, but instead to encode some sense of relatable semantics of hidden states as a whole.
Since some hidden states naturally bear little stand-alone representational power, we expect their semantics to be highly ambiguous as these findings indicate.
With an appropriate visual encoding, this ambiguity will pop out in a visualization and help lead users to the same conclusion about the semantics of these hidden states.


\section{Summary}\label{summary}
We introduce a technique for representing hidden states called Predictive Semantic Encodings (PSE).
PSEs are a formulation that expresses the meaning of hidden states as a whole in relation to the task inputs or outputs.
We propose a visual encoding for this technique that is comparable across differing kinds of hidden states, making it a powerful tool in visualizing the interaction between components of RNNs.

Although this idea is simple, its key contribution is allowing for visualizations to show the high level intuitions captured within hidden states.
Moreover, its compact form make rendering many hidden states at once possible, opening avenues for visualizations to better investigate the internal details of these complex models.

We use this technique to perform a preliminary investigation into a realistic task and dataset; that of a language model.
This visual encoding brings to light previously un-visualized aspects of the RNN model behaviour.
Moreover, it highlights various areas of interest in the model such as where significant shifts occur in hidden state semantics.
The evaluation concludes by showing how the PSE is equivalent to the model classifier, as well as by using it to explain some of the complexity represented across the hidden states of the RNN itself.



\acknowledgments{
We thank the anonymous reviewers who's comments and suggestions greatly helped to improve and clarify this manuscript.}

\bibliographystyle{abbrv-doi}

\bibliography{template}

\section*{Appendix: Long Short-Term Memory}\label{appendix:lstm}
The recurrence function of the LSTM follows.
We use subscripts $t$ to denote the timestep through the sequential data, and superscripts $u$ to denoted the stacked layers (units) of the recurrent function.

\begin{align*}
\mathbf{f}_t^u &= \sigma (W_f [\mathbf{h}_t^{u-1} , \mathbf{h}_{t-1}^u] + \mathbf{b}_f) & \text{Forget Gate} \\
\mathbf{i}_t^u &= \sigma (W_i [\mathbf{h}_t^{u-1} , \mathbf{h}_{t-1}^u] + \mathbf{b}_i) & \text{Remember Gate} \\
\mathbf{o}_t^u &= \sigma (W_o [\mathbf{h}_t^{u-1} , \mathbf{h}_{t-1}^u] + \mathbf{b}_o) & \text{Output Gate} \\
\mathbf{\tilde{c}}_t^u &= \tanh (W_c [\mathbf{h}_t^{u-1} , \mathbf{h}_{t-1}^u] + \mathbf{b}_c) & \text{Cell Input} \\
\mathbf{l}_t^u &= \mathbf{f}_t^u \circ \mathbf{c}_{t-1}^u & \text{Long-term Memory} \\
\mathbf{s}_t^u &= \mathbf{i}_t^u \circ \mathbf{\tilde{c}}_t^u & \text{Short-term Memory} \\
\mathbf{c}_t^u &= \mathbf{l}_t^u + \mathbf{s}_t^u & \text{Cell State} \\
\mathbf{h}_t^u &= \mathbf{o}_t^u \circ \tanh(\mathbf{c}_t^u) & \text{Output State}
\end{align*}

Where $W_*$ represents a matrix of the size $N x 2N$ and $\mathbf{b}_*$ a bias vector $N x 1$.

\end{document}